\documentclass{article}

\usepackage{PRIMEarxiv}

\usepackage[utf8]{inputenc} % allow utf-8 input
\usepackage[T1]{fontenc}    % use 8-bit T1 fonts
\usepackage{hyperref}       % hyperlinks
\usepackage{url}            % simple URL typesetting
\usepackage{booktabs}       % professional-quality tables
\usepackage{amsfonts}       % blackboard math symbols
\usepackage{nicefrac}       % compact symbols for 1/2, etc.
\usepackage{microtype}      % microtypography
\usepackage{lipsum}
\usepackage{fancyhdr}       % header
\usepackage{graphicx}       % graphics
\usepackage{amsmath}
\graphicspath{{media/}}     % organize your images and other figures under media/ folder

\title{Explainable Machine Learning System for Predicting\\ Chronic Kidney Disease in High-Risk Cardiovascular Patients
}

\author{
  Nantika Nguycharoen \\
  Faculty of Engineering and Technology \\
  Liverpool John Moores University \\
  Liverpool, UK\\
  \texttt{n.nguycharoen@2023.ljmu.ac.uk}, \\ \texttt{nantika.ngr@gmail.com} \\
  }

\begin{document}
\maketitle
\begin{abstract}
As the global population ages, the prevalence of chronic conditions such as Chronic Kidney Disease (CKD) is increasing. CKD often remains asymptomatic until it reaches advanced stages, which creates significant burdens for both the healthcare system and patients' quality of life. This research addresses the need for early detection by developing an explainable machine learning system to predict CKD in patients with cardiovascular risks using medical history and laboratory data.

The Random Forest model was selected as this study's primary machine learning model because of its highest sensitivity rate of 88.2\%. The emphasis on sensitivity or minimizing false negatives is crucial, given the model's intended use in the screening process. The study introduces a novel comprehensive explainability framework that extends beyond traditional feature importance methods. It includes global and local interpretations, bias inspection, biomedical relevance, and safety assessments. The system identified crucial predictive features, including the usage of diabetic medication, initial estimated glomerular filtration rate (eGFR) value, and the usage of Angiotensin-converting enzyme inhibitors/Angiotensin receptor blockers (ACEI/ARB) medication. It also provided insights into the model's decision-making through counterfactual explanations. The system observed no significant gender bias and identified certain biases concerning initial eGFR values and CKD predictions. The model's logic, extracted through scoped rules, was aligned with established medical literature and showed no conflicts. Additionally, the safety assessment results indicated that potentially dangerous cases were managed safely. 

The developed system not only enhances the explainability, reliability, and accountability of the predictive model, but also supports its potential adoption in healthcare environments and adherence to evolving regulatory standards. The framework established here also holds promise for application across various healthcare machine learning contexts.
\end{abstract}

% keywords can be removed

\section{Introduction}
The aging global population is leading to an increased prevalence of chronic diseases, subsequently overwhelming healthcare facilities with a growing number of patients. Among these, Chronic Kidney Disease (CKD) stands out as particularly prevalent, currently affecting over 800 million people worldwide, which represents more than 10\% of the global population \cite{kovesdy_epidemiology_2022}. With rising life expectancy, these figures are anticipated to increase further. CKD is a leading cause of death in high-income countries, closely linked with heart disease and stroke, the foremost global causes of mortality\cite{WHO2020,CDC}.

CKD involves a gradual decline in kidney function, caused by factors such as diabetes (38\%), hypertension (27\%), and glomerulonephritis (15\%), with additional risks posed by heart disease, obesity, family history, and advancing age \cite{CDC}. Its often asymptomatic early stages and non-specific symptoms complicate timely diagnosis. The challenge of timely CKD diagnosis is compounded by the sheer volume of patients in the healthcare system, leading to delays in recognizing the disease. However, the increasing adoption of electronic health systems offers a pathway to automated CKD detection solutions.

Despite significant research on applying machine learning for CKD prediction, facilitated by the growing availability of open-source data and technological advancements, real-world clinical applications remain scarce. Healthcare's sensitive nature, dealing directly with human lives, categorizes it as a high-risk area. The opaque decision-making processes of machine learning systems further hinder their acceptance. Although recent studies have started to focus on explainability, they often limit themselves to analyzing feature importance. Moreover, healthcare applications must adhere to stringent standards, with emerging AI regulations scrutinizing their risk, safety, transparency, and other aspects, posing additional barriers to the adoption of machine learning in healthcare.

In order to bridge the existing gap, this study develops an explainable machine learning system for CKD prediction anchored by two main components: a high-sensitivity machine learning model utilizing medical history and laboratory values for prediction and an explainability framework designed to foster real-world acceptance.

Developing an explainable machine learning system capable of predicting CKD could streamline CKD detection by integrating it with electronic health record systems for screening. Such a system promises to facilitate early detection and treatment, improve patient outcomes, reduce the workload on medical staff, and increase system acceptance through comprehensive explanations.

%\section{Related Works}
\section{Methods}
This study used two main components to build an explainable machine learning system for predicting CKD. The first component was a machine learning model to predict CKD, while the second component was an explainable system. These two parts can be described below.

\subsection{Machine Learning Model}
The study developed a machine learning model using the dataset from the retrospective study by Al-Shamsi et al.\cite{al-shamsi}, which studied CKD incidence and risk factors in patients with or at high-risk cardiovascular disease. In this dataset, patients initially did not have CKD and were monitored for the progression to CKD. This strategy aligns with real-world clinical practice, which aims to detect CKD at its earliest stage. The data was obtained from the outpatient electronic medical records of Tawam Hospital, located in Al-Ain City, UAE.

The dataset contains both numerical and categorical features. The numerical features consist of age, body mass index (BMI), systolic blood pressure (SBP), diastolic blood pressure (DBP), cholesterol level (Chol), triglyceride level (TG), Hemoglobin A1C (HbA1c), creatinine level (Cr), and estimated glomerular filtration rate (eGFR). Meanwhile, the categorical features comprise gender and the presence or absence of a history of various illnesses, including coronary heart disease (CHD), diabetes mellitus (DM), vascular disease, hypertension (HT), dyslipidemia (DLP), smoking, and obesity. Additionally, the categorical features include a history of drug use, including Angiotensin-converting-enzyme Inhibitors or Angiotensin Receptor Blockers medicine (ACEI/ARB), dyslipidemia medicine, diabetic medicine, and hypertension medicine. The target variable is whether the patient has CKD stages 3 to 5.

During the data preprocessing step, it was discovered that two features in the dataset had missing values. The first feature was Hemoglobin A1C (HbA1c), which was imputed with the mean value of HbA1c of their respective group - diabetic or non-diabetic. The second feature was the triglyceride level, which was also imputed with the mean triglyceride level of their respective group - dyslipidemia or non-dyslipidemia. The numerical data was then rescaled using normalization.

The dataset used in this work is relatively small, with only 491 entries, and it suffers from an imbalanced class problem. The majority class is negative (no CKD), accounting for 88.6\% of the dataset, while the positive class (CKD) only accounts for 11.4\%. Due to this problem, the Synthetic Minority Over-sampling Technique for Nominal and Continuous (SMOTE-NC) was used in this study.

Five machine learning algorithms were experimented with and compared: Logistic Regression, Decision Tree, Random Forest, XGBoost, and ANN, to ensure a range of models from simple to complex. The study also performed hyperparameter tuning during the model training process. The best-performing model was then proceeded to the next step of building an explainable system.

\subsection{Explainable System}
The study proposed a novel system for interpreting and explaining the outcome model. The explainable system was designed into five parts: global interpretation, local interpretation, bias inspection, biomedical relevance, and safety assessment. Figure 1 shows the overview of the explainable system.

\begin{figure}[ht]
\centering
\includegraphics[width=0.5\textwidth] {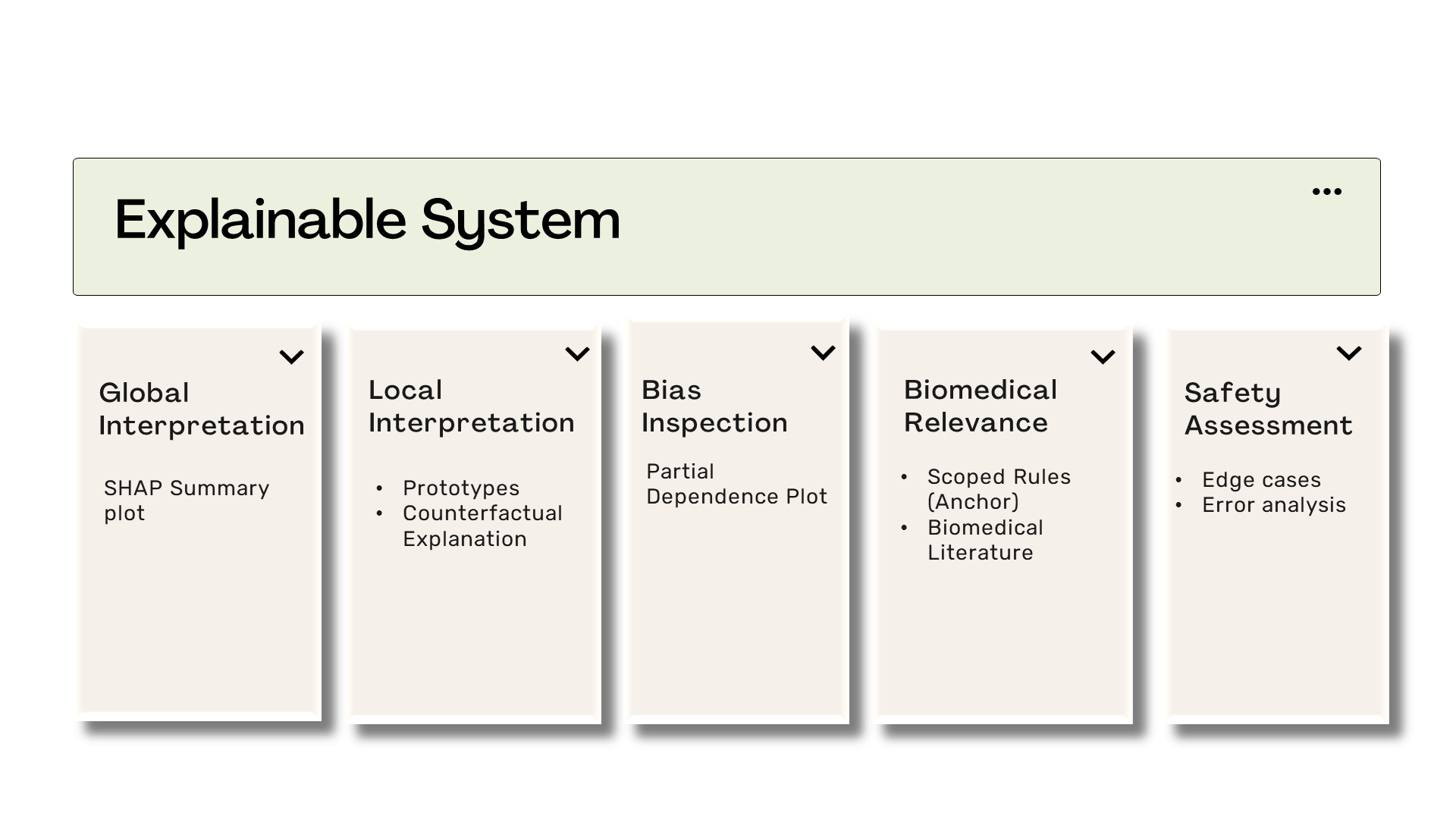}
\caption{The explainable system}
\label{fig1}
\end{figure}

\subsubsection{Global interpretation} 
The system first focused on overviewing the model globally. The SHAP summary plot was selected as it displays the overall importance of features and reveals whether a feature has a positive or negative impact on predictions. The Shapley value is a concept that originated from coalitional game theory. It is used to estimate a feature's contribution to a model, which can be measured as the increase in probability for classification problems. In order to calculate the Shapley value of each feature, the average marginal contribution of the feature across all possible subsets of features is taken \cite{Masis2021}.

\subsubsection{Local Interpretation} 
The study then moved to exploring the model locally. To explain the model locally, prototypes were initially identified as key instances representing all samples. Each prototype was further analyzed through counterfactual explanations. This process allowed the study to explain and interpret the model locally while ensuring thorough coverage of the data distribution. 

The prototype selection method used in this study was derived from the work by Bien and Tibshirani \cite{Bien2011}. The prototype selection aims to identify the most essential subset of samples that can provide a distilled or condensed perspective of a given dataset. The concept of counterfactuals is derived from human reasoning. It involves identifying the smallest change required to produce the opposite outcome. The algorithmic approach to finding counterfactual instances involves identifying the closest points with different outcomes. The study chose to use Google's What-If Tool (WIT) as a tool to find the counterfactual instances. The WIT displays an interactive dashboard that shows data points and their counterfactuals.

\subsubsection{Bias Inspection} 
So as to assess any potential bias, the focus was on the patient's gender and initial eGFR value. Although gender can be considered a biological difference, it still requires careful inspection. The initial eGFR values may have some bias as lower values are closer to the cutoff eGFR of CKD. 
A partial dependence plot (PDP) was used to reflect the features' effect on prediction. This method was chosen because it can reflect the relationship between a feature of interest and a target variable. 

The partial dependence marginalizes the model output across the uninterested feature distributions to ensure that it only shows the relationship between the interested feature and the predicted outcome \cite{Molnar2022}. The partial dependence plot is the plot between the values of the feature of interest on the x-axis and the average prediction on the y-axis. The limitation of PDP is that it assumes a feature of interest is independent of other features.

\subsubsection{Biomedical Relevance} 
In an effort to evaluate the relevance of the system's logic to biomedicine, the study utilized anchor explanations, also known as scoped rules, to extract information from the model. A scoped rule is a simple if-then rule that explains the prediction of an instance \cite{Molnar2022}. The extracted rules were then cross-checked against existing biomedical knowledge and research. This method was chosen as it allows for the extraction of information in a clear and logical manner.

\subsubsection{Safety Assessment}  
Two approaches were utilized to assess safety. The first approach involved creating edge cases as safety test cases. These edge cases were designed to test the model's behavior in extreme feature values or potentially dangerous situations, drawing inspiration from computer programming. The second approach examined instances where the model provided incorrect predictions to identify root causes and determine the extent of the error.

\section{Results}
Among the five machine learning models used to experiment in building a predictive model for CKD, the random forest model achieved the highest sensitivity of 0.882. The system was prioritized for sensitivity because it was designed to be used as a medical screening tool. 

Compared to another study that employed the same dataset for training a machine learning model, Chicco et al.\cite{Chicco2021} also discovered that the random forest model was the best model. However, this study's random forest model outperformed the one used in Chicco et al.\cite{Chicco2021}'s study in terms of sensitivity (0.882 sensitivity, 0.641 specificity, 0.840 ROC-AUC in this study versus 0.793 sensitivity, 0.852 specificity, 0.885 ROC-AUC in Chicco et al.\cite{Chicco2021}'s study). Nevertheless, the other metrics were comparatively lower, indicating that there were trade-offs between sensitivity and other metrics. After finalizing the main predictive model, the next step was to develop an explainable system.

\subsection{Global Interpretation}
\begin{figure}[ht]
  \begin{minipage}[b]{0.48\linewidth}
    \centering
    \includegraphics[width=\linewidth]{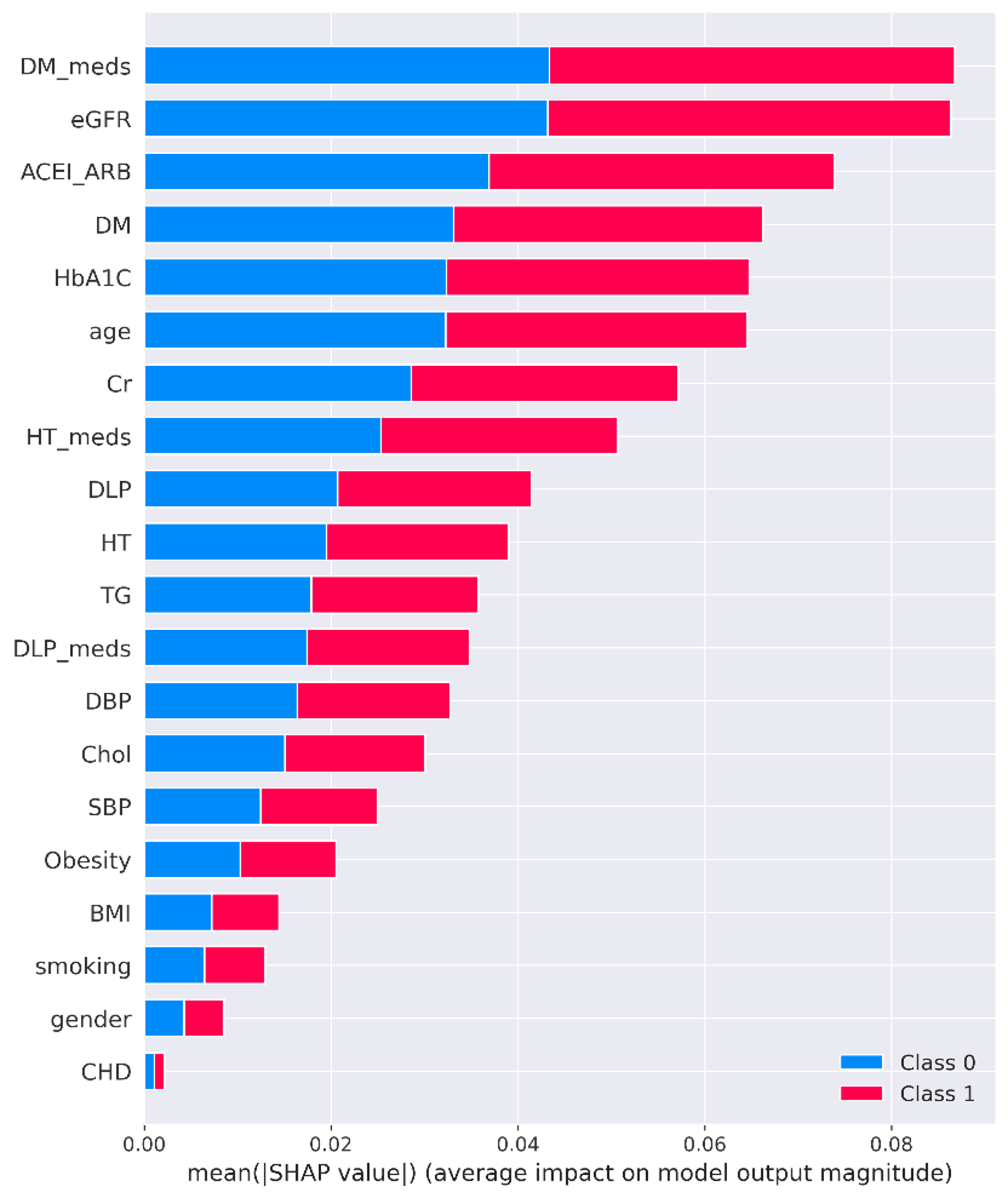}
    \caption{The SHAP summary plot for the random \\forest model}
    \label{fig2}
  \end{minipage}
  \hfill
  \begin{minipage}[b]{0.5\linewidth}
    \centering
    \includegraphics[width=\linewidth]{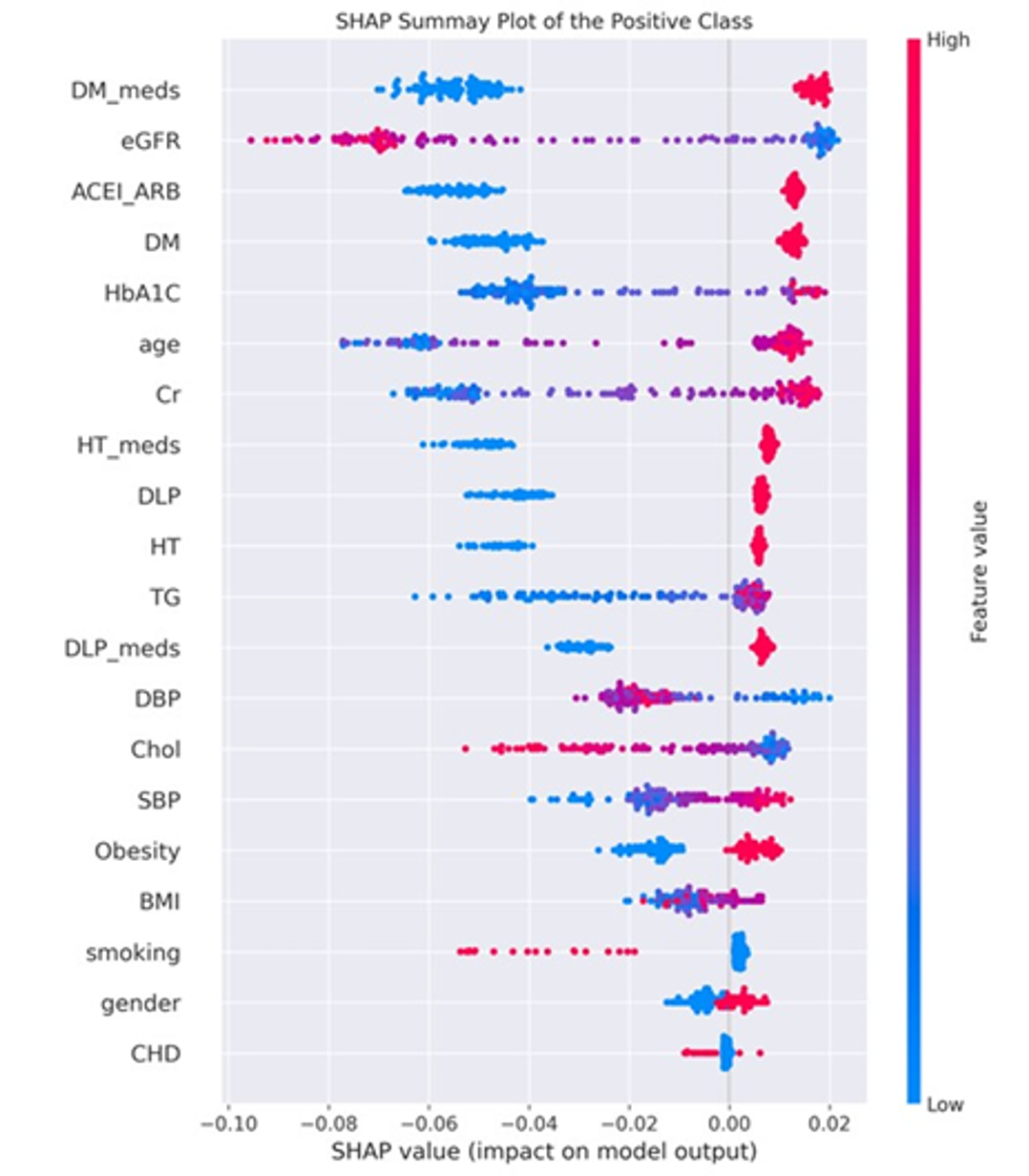}
    \caption{The SHAP summary plot shows the distribution of SHAP values for predictions of CKD.}
    \label{fig3}
  \end{minipage}
\end{figure}

The SHAP summary plot in Figure 2 shows the overall features’ importance of each feature. Each bar represents the mean absolute SHAP value for a feature. The features are ranked by the sum of SHAP value magnitudes over all samples. The top five most important features are diabetic medication (DM\_meds), eGFR, ACEI/ARB medication (ACEI\_ARB), Diabetes (DM), and HbA1C. 
Figure 3 provides a more detailed analysis of the SHAP values. The high values of DM\_meds, ACEI\_ARB, DM, HbA1c, and age contribute more to CKD prediction. However, the lower eGFR value contributes more to CKD prediction.

\subsection{Local Interpretation}
The study identified prototypes for further analysis with counterfactuals. It is noted that these prototypes all belonged to the negative class (no CKD). This is because the dataset is highly imbalanced, with the majority class being a negative class. The values in some categorical features, such as CHD, vascular disease, and smoking, were the same among the ten prototypes. This is also because the ratios of samples having that feature were very low. For numerical features, while age and Cr exhibited a wide range of values across the prototypes, other numeric features such as Cholesterol and HbA1c were more limited in range. The prototypes and their feature values are shown in Table 1.

\begin{table}[ht]
\centering
\resizebox{\textwidth}{!}{
\begin{tabular}{|l|l|l|l|l|l|l|l|l|l|l|}
\hline
Feature          & prototype 1 & Prototype 2 & Prototype 3 & Prototype 4 & Prototype 5 & Prototype 6 & Prototype 7 & Prototype 8 & Prototype 9 & Prototype 10 \\ \hline
gender           & Woman       & Woman       & Woman       & Woman       & Man         & Woman       & Woman       & Man         & Woman       & Man          \\ \hline
age              & 28          & 57          & 42          & 48          & 30          & 74          & 68          & 26          & 43          & 36           \\ \hline
DM               & no          & no          & no          & no          & no          & yes         & no          & no          & no          & no           \\ \hline
CHD              & no          & no          & no          & no          & no          & no          & no          & no          & no          & no           \\ \hline
vascular disease & no          & no          & no          & no          & no          & no          & no          & no          & no          & no           \\ \hline
smoking          & no          & no          & no          & no          & no          & no          & no          & no          & no          & no           \\ \hline
HT               & no          & no          & no          & yes         & no          & yes         & yes         & no          & no          & no           \\ \hline
DLP              & no          & yes         & no          & no          & no          & yes         & yes         & no          & no          & no           \\ \hline
Obesity          & no          & yes         & no          & yes         & no          & no          & yes         & no          & yes         & yes          \\ \hline
DLP\_meds        & no          & yes         & no          & no          & no          & yes         & yes         & no          & no          & no           \\ \hline
DM\_meds         & no          & no          & no          & no          & no          & no          & no          & no          & no          & no           \\ \hline
HT\_meds         & no          & no          & no          & yes         & no          & yes         & yes         & no          & no          & no           \\ \hline
ACEI\_ARB        & no          & no          & no          & no          & no          & no          & no          & no          & no          & no           \\ \hline
Chol             & 5.34        & 4.3         & 3.9         & 5.6         & 4.94        & 4.7         & 4.7         & 3.9         & 4.8         & 4.8          \\ \hline
TG               & 1.18        & 1.52        & 1.45        & 0.89        & 1.3         & 0.99        & 1.44        & 0.5         & 0.6         & 1.01         \\ \hline
HbA1C            & 5.2         & 5.6         & 5.9         & 5.7         & 5.68        & 6.54        & 5.5         & 5.6         & 5.2         & 5.5          \\ \hline
Cr               & 55          & 59          & 50          & 40          & 78          & 54          & 49          & 59          & 37          & 79           \\ \hline
eGFR             & 122.9783    & 98.02245    & 115.0101    & 118.6629    & 115.1651    & 89.5603     & 96.45019    & 132.8479    & 126.098     & 109.8355     \\ \hline
SBP              & 101         & 130         & 115         & 133         & 112         & 146         & 141         & 122         & 122         & 128          \\ \hline
DBP              & 64          & 79          & 73          & 82          & 74          & 78          & 80          & 78          & 76          & 81           \\ \hline
BMI              & 23.91784    & 31.13093    & 28.23838    & 34.82935    & 21.19274    & 26.89767    & 41.55125    & 22.62518    & 29.93069    & 30.66406     \\ \hline
Label            & no          & no          & no          & no          & no          & no          & no          & no          & no          & no           \\ \hline
\end{tabular}
}
\caption{The prototypes of the dataset and their feature values}
\label{table1}
\end{table}

\begin{table}[ht]
  \begin{minipage}{0.5\textwidth}
    \centering
    \begin{tabular}{|l|l|l|}
\hline
type              & Reference & Counterfactual \\ \hline
gender            & woman     & woman          \\ \hline
age               & 57        & 62             \\ \hline
DM                & no        & no             \\ \hline
CHD               & no        & no             \\ \hline
Vascular\_disease & no        & no             \\ \hline
smoking           & no        & no             \\ \hline
HT                & no        & yes            \\ \hline
DLP               & yes       & yes            \\ \hline
Obesity           & yes       & yes            \\ \hline
DLP\_meds         & yes       & yes            \\ \hline
DM\_meds          & no        & no             \\ \hline
HT\_meds          & no        & yes            \\ \hline
ACEI\_ARB         & no        & yes            \\ \hline
Chol              & 4.3       & 5.1            \\ \hline
TG                & 1.52      & 1.01           \\ \hline
HbA1C             & 5.6       & 5.8            \\ \hline
Cr                & 59        & 63             \\ \hline
eGFR              & 98.02245  & 91.19419       \\ \hline
SBP               & 130       & 122            \\ \hline
DBP               & 79        & 68             \\ \hline
BMI               & 31.13093  & 29.5369        \\ \hline
Prediction        & No CKD    & CKD            \\ \hline
\end{tabular}

    \caption{The example of the counterfactual explanation}
  \end{minipage}%
  \hfill
  \begin{minipage}{0.5\textwidth}
    \centering
    \begin{tabular}{|l|l|l|}
\hline
type              & Reference & Counterfactual \\ \hline
gender            & woman     & woman          \\ \hline
age               & 48        & 51             \\ \hline
DM                & no        & yes            \\ \hline
CHD               & no        & no             \\ \hline
Vascular\_disease & no        & no             \\ \hline
smoking           & no        & no             \\ \hline
HT                & yes       & yes            \\ \hline
DLP               & no        & no             \\ \hline
Obesity           & yes       & yes            \\ \hline
DLP\_meds         & no        & no             \\ \hline
DM\_meds          & no        & yes            \\ \hline
HT\_meds          & yes       & yes            \\ \hline
ACEI\_ARB         & no        & no             \\ \hline
Chol              & 5.6       & 3.2            \\ \hline
TG                & 0.89      & 1.92           \\ \hline
HbA1C             & 5.7       & 6.6            \\ \hline
Cr                & 40        & 61             \\ \hline
eGFR              & 118.6629  & 101.1269       \\ \hline
SBP               & 133       & 140            \\ \hline
DBP               & 82        & 80             \\ \hline
BMI               & 34.82935  & 39.79849       \\ \hline
Prediction        & No CKD    & CKD            \\ \hline
    \end{tabular}
    \caption{The example of the counterfactual explanation}
  \end{minipage}
\end{table}

Each prototype was used as a key instance to find its corresponding counterfactual instance. Below are some intriguing prototypes and counterfactual pairs. 

From Table 2, both patients were at almost similar ages, with obesity, dyslipidemia, and receiving treatment for dyslipidemia. However, the patient who developed CKD had hypertension and was being treated with hypertensive medication and ACEI/ARB medication. This suggested that hypertension and hypertensive-related conditions increase the risk of developing CKD.

From Table 3, it can be observed that both patients were women at almost similar ages and had hypertension. The most significant difference was that the patient who developed CKD had diabetes and was receiving diabetic medicine. This highlighted the significant role of diabetes in increasing the risk of CKD.

\subsection{Bias Inspection}
\begin{figure}[ht]
  \begin{minipage}{0.5\linewidth}
    \centering
    \includegraphics[height=4cm,keepaspectratio]{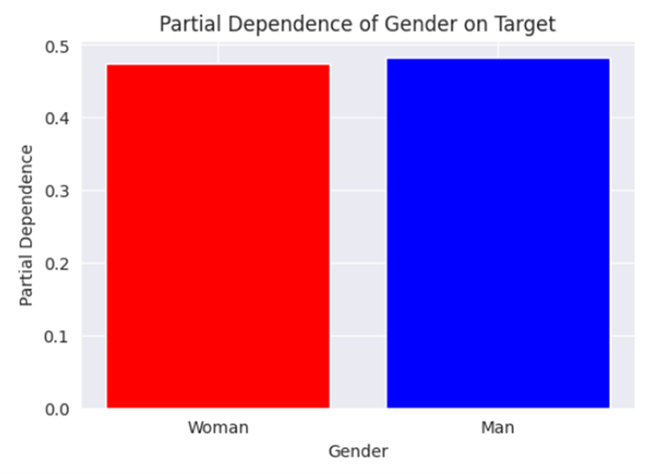}
    \caption{The partial dependence plot of gender\\ 
    on the target}
    \label{fig4}
  \end{minipage}
  \hfill
  \begin{minipage}{0.5\linewidth}
    \centering
    \includegraphics[height=4cm,keepaspectratio]{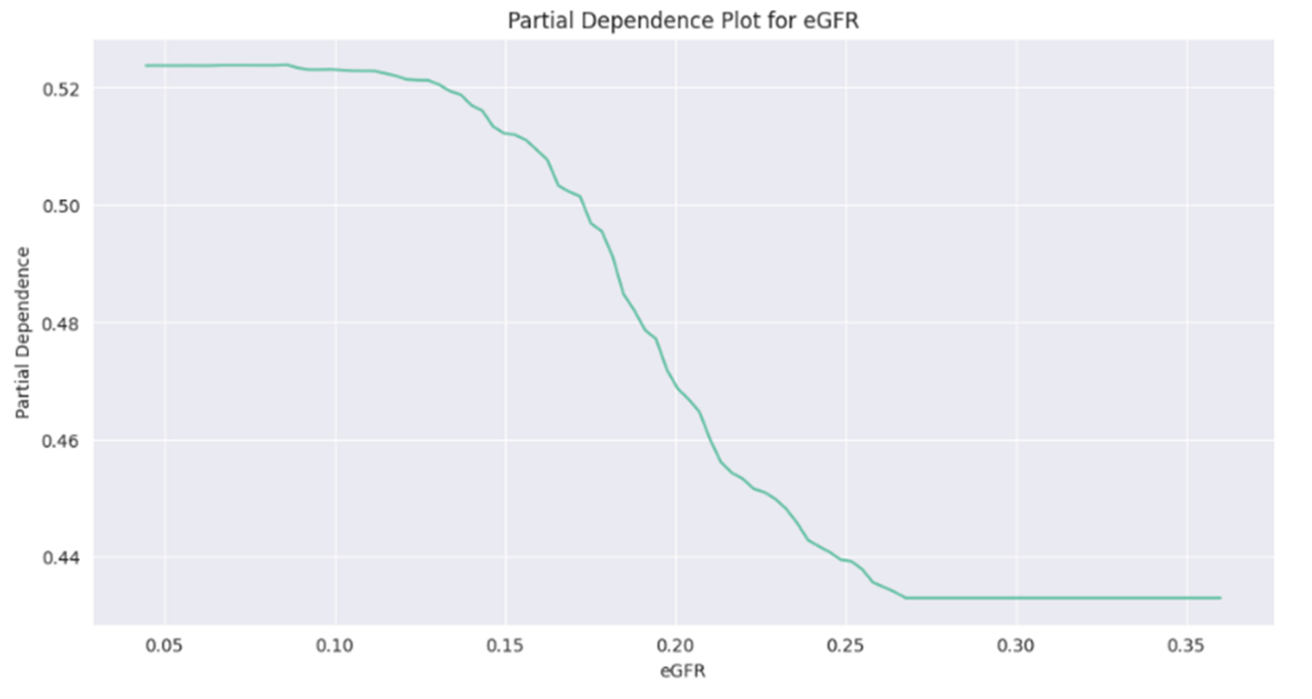}
    \caption{The partial dependence plot of eGFR on the target}
    \label{fig5}
  \end{minipage}
\end{figure}
To explore gender bias, the partial dependence plot of the gender and the target was plotted as shown in Figure 4. It is observed that there was a slight difference in partial dependence values between women and men. This means that after averaging out the effects of all other features in the model, the average predicted probability of having CKD for women was slightly lower than for men. So, there was no obvious gender bias.

Because the eGFR is involved in the criteria for diagnosis of CKD. The study decided to investigate the bias of the initial eGFR to the CKD prediction. The partial dependence plot, as shown in Figure 5, illustrates that lower eGFR values have a higher average predicted probability of having CKD than the higher eGFR values. Especially when the eGFR increases from 86 mL/min/1.73 $\text{m}^2$ (corresponding to 0.14 of the scaled eGFR in the graph) to 106 mL/min/1.73 $\text{m}^2$ (corresponding to 0.25 of the scaled eGFR in the graph), the average predicted probability decreases significantly. However, at very low and very high eGFR values, the relationship is less pronounced, indicating that the probability of CKD is less sensitive to changes in eGFR within these ranges. This implied that there was some degree of bias between the initial eGFR values and the prediction of CKD.

\subsection{Biomedical Relevance}
The study used scoped rules (anchor) to acquire knowledge from the model and cross-check it with current medical knowledge. The interesting scoped rules were listed and analyzed with current medical knowledge below.

Rule No. 1: If eGFR is less than or equal to 87.74 and DM is yes, predict class CKD with precision of 94.95\% and coverage of 40.23\%. It can be noticed from this rule that patients who have low initial eGFR with diabetes conjunction factor are likely to develop CKD. This aligned with the research from Zelnick et al.\cite{Zelnick2017}, who found that diabetes is significantly related to albuminuria and decreased GFR regardless of demographics and hypertension. Diabetes is also the most common cause of CKD \cite{Pyram2012}.

Rule No. 2: If HbA1C is greater than 6.76 and CHD is Yes, then predict class CKD with precision of 95.81\% and coverage of 4.78\%. This rule can be interpreted that patients with high HbA1C levels (which indicates diabetes) and coronary heart disease (CHD) are likely to develop CKD. The coverage percentage of this rule was low because very few patients in this dataset had CHD. The rule aligned with the current biomedical knowledge that coronary heart disease is a risk factor for CKD and vice versa\cite{Weiner2006, CDC}. So, the rule summarized that patients with two risk factors, diabetes and CHD, are highly susceptible to developing CKD.

Rule No. 3: If HT\_meds is no and eGFR > 100.5 and ACEI\_ARB is no, then predict class no CKD with a precision of 97.50\% and coverage of 12.92\%. This rule stated that patients with good baseline kidney function and blood pressure that is not high enough for antihypertensive or ACEI/ARB treatment are unlikely to develop CKD. The rule is logical because hypertension is the second leading cause of CKD\cite{CDC}. Hypertension and CKD are closely related. Sustained hypertension can worsen kidney function, and declining kidney function can lead to poor blood pressure control\cite{Ku2019}.

Rule No. 4: If age $\leq 51$ and eGFR $> 100.5$ and DM\_meds is no, then predict class no CKD with precision: 100.00\% and coverage: 15.26\%. According to this rule, individuals who are 51 years old or younger with good baseline kidney function and good blood sugar levels that do not require diabetic treatment are at a lower risk of developing CKD. The prevalence of CKD is higher in individuals aged 65 years or older compared to other age groups\cite{CDC}. Vascular aging can cause the vessel wall of the kidney to lose elasticity and compliance, leading to Chronic Kidney Disease. However, in the absence of progressive cardiovascular disease, the decline in renal function due to normal aging is slow and does not appear to be of significant clinical implication\cite{Abdelhafiz2010}. So, it was reasonable to assume that patients without two risk factors, i.e., older age and diabetes, are less likely to develop CKD.

Rule No. 5: If ACEI\_ARB is no, and DLP is no, and HbA1C $\leq 5.88$, then predict class no CKD with precision: 100.00\% and coverage: 9.15\%. This rule stated that patients with prediabetes or no diabetes (corresponding to HbA1C $\leq 5.88$) and without dyslipidemia and without condition that needs to be treated with ACEI/ARB are less likely to develop CKD. Prediabetes is associated with a modest increase in chronic kidney disease risk, but this association has not been confirmed conclusively\cite{Echouffo2016,Hennessey2020}. Dyslipidemia is associated with an increased risk of CKD\cite{Liu2013,Liang2020}. While ACEI and ARBs are anti-hypertensive drugs that are effective in reducing proteinuria and preserving kidney function, in some cases, they can cause adverse effects of declining renal function\cite{Momoniat2019}. As a result, patients without these three factors, i.e., Diabetes, Dyslipidemia, and taking ACEI/ARB medications, are less likely to develop CKD.

\subsection{Safety Assessment}
The study conducted a safety assessment of the model by designing test cases that evaluate the model's safety in potentially harmful and extreme conditions. The study also examined the extent of errors in cases where the model made false predictions. Below are some interesting cases that the study explored during the safety assessment.

Edge case 1 (healthy case): Female 25 years old without any medical comorbidity ({'gender': 0, 'age': 25, 'DM': 0, 'CHD': 0, 'Vascular\_disease': 0, 'smoking': 0, 'HT': 0, 'DLP': 0, 'Obesity': 0, 'DLP\_meds': 0, 'DM\_meds': 0, 'HT\_meds': 0, 'ACEI\_ARB': 0, 'Chol': 3.1, 'TG': 0.68, 'HbA1C': 5, 'Cr': 61, 'eGFR': 123, 'SBP': 120, 'DBP': 80, 'BMI': 19}). Even though the model was designed for patients with CVD risk, this case was examined as a sanity check. The model predicted no CKD with a probability of no CKD equal to 0.83. This case was considered a pass.

Edge case 2 (patients with all comorbidities): Male 80 years old with DM, HT, DLP, CHD, vascular disease, smoking, obesity, and receiving treatments ({'gender': 1, 'age': 80, 'DM': 1, 'CHD': 1, 'Vascular\_disease': 1, 'smoking': 1, 'HT': 1, 'DLP': 1, 'Obesity': 1, 'DLP\_meds': 1, 'DM\_meds': 1, 'HT\_meds': 1, 'ACEI\_ARB': 1, 'Chol': 9.05, 'TG': 2.26, 'HbA1C': 7.5, 'Cr': 106, 'eGFR': 61, 'SBP': 180, 'DBP': 100, 'BMI': 30}). The model predicted CKD with a probability of 0.76. This case was considered a pass.

Edge case 3 (patients with all comorbidities but untreated): Male 80 years old with DM, HT, DLP, CHD, vascular disease, smoking, obesity, and not receiving any medication ({'gender': 1, 'age': 80, 'DM': 1, 'CHD': 1, 'Vascular\_disease': 1, 'smoking': 1, 'HT': 1, 'DLP': 1, 'Obesity': 1, 'DLP\_meds': 0, 'DM\_meds': 0, 'HT\_meds': 0, 'ACEI\_ARB': 0, 'Chol': 9.05, 'TG': 2.26, 'HbA1C': 7.5, 'Cr': 106, 'eGFR': 61, 'SBP': 180, 'DBP': 100, 'BMI': 30}). The model predicted CKD with a probability of 0.55. Even though receiving medications can be viewed as relieving factors, the model considered medication factors as risk factors, similar to the diseases themselves. This aligned with information from the SHAP summary plot. So, the probability was decreased compared to the second edge case. 

Edge case 4 (the false negative case): Male 75 years old without comorbidity, but having low initial eGFR ({'gender': 1, 'age': 75, 'DM': 0, 'CHD': 0, 'Vascular\_disease': 0, 'smoking': 0, 'HT': 0, 'DLP': 0, 'Obesity': 0, 'DLP\_meds': 0, 'DM\_meds': 0, 'HT\_meds': 0, 'ACEI\_ARB': 0, 'Chol': 5.78, 'TG': 0.88, 'HbA1C': 6.21, 'Cr': 100.77, 'eGFR': 61.82, 'SBP': 134, 'DBP': 74, 'BMI': 21}). The ground truth of this case was CKD, but the model predicted no CKD with a probability ratio of 0.67:0.33 (no CKD: CKD). It can be assumed that the model made an incorrect prediction because the prominent risk factor in this case was only the low eGFR, while many risk factors were absent. The partial dependence value at this eGFR level was approximately 0.525, as displayed in Figure 5. When combined with the absence of other risk factors, the model then predicted a lower probability of CKD.

Edge case 5 (the false negative case): Male 69 years old with smoking, HT, receiving treatment of HT, slightly elevated Cr level, and low initial eGFR ({'gender': 1, 'age': 69, 'DM': 0, 'CHD': 0, 'Vascular\_disease': 0, 'smoking': 1, 'HT': 1, 'DLP': 0, 'Obesity': 0, 'DLP\_meds': 0, 'DM\_meds': 0, 'HT\_meds': 1, 'ACEI\_ARB': 0, 'Chol': 5.78, 'TG': 1.42, 'HbA1C': 5.55, 'Cr': 107.79, 'eGFR': 60.05, 'SBP': 166, 'DBP': 92, 'BMI': 18}). The ground truth of this case was CKD, but the model predicted no CKD with a probability ratio of 0.55:0.45 (no CKD: CKD). It was observed that in this case, the patient’s initial eGFR of this case was very close to the cutoff point for CKD diagnosis and had hypertension as a risk factor. However, the model missed the prediction by a small margin. This could be because the model did not account for smoking as a significant risk factor. This information was consistent with the SHAP summary plot above.

\section{Discussion}
The study's limitation lies in the dataset's regional specificity (United Arab Emirates) and its small size, especially regarding CKD cases, which affects generalizability. Despite its limitations, the strength of this dataset lies in its ability to track patient data from no CKD to the development of CKD, which is not available in any other public dataset in this field. This longitudinal data complies with the practical CKD management strategy for prevention and early detection. The other positive aspect of the dataset is its focus on high-risk cardiovascular disease patients. This helps to narrow the search for high-risk CKD patients. Therefore, future research should aim at larger, more diverse populations to improve the model’s generalizability and performance.

The machine learning model used in this work showed higher sensitivity, which is comparable to another previous study. However, there is still scope for improvement, particularly in terms of its performance on other metrics. This issue is also associated with the dataset, which is small and has a low number of positive classes. Hence, future research conducted on a larger population is recommended.

The novel explainable system developed in this study was designed specifically for predicting chronic kidney disease (CKD) using a particular dataset. However, the methods and ideas used in this research show potential for use in other machine learning applications in healthcare. This could broaden the impact of the study beyond CKD prediction and benefit other areas of healthcare.

\section{Conclusion}
This research addressed the growing challenge of chronic kidney disease (CKD) among an aging global population by developing an explainable machine learning system for predicting CKD that leveraging patients' medical histories and laboratory data. 

The developed random forest model, chosen for its high sensitivity of 0.882, reflects the critical importance of minimizing false negatives in CKD screening despite trade-offs with specificity and accuracy. This approach aligns with the imperative in healthcare to avoid overlooking potential CKD cases.

Central to this study is the novel explainability system designed to demystify machine learning predictions for CKD, enhancing the model's interpretability and reliability. The system's explainable architecture comprises five distinct components, i.e., global interpretation, local interpretation, bias inspection, biomedical relevance, and safety assessment. The system was designed to bridge the gap between complex machine learning algorithms and practical healthcare applications, thereby facilitating broader acceptance and compliance with regulatory standards.

Key findings from the system include identifying influential features such as diabetic medication usage, estimated Glomerular Filtration Rate (eGFR), and ACEI/ARB medication usage through global interpretation via SHAP summary plots. Local interpretation, where representative examples (prototypes) were selected to explain the model's decision through counterfactual explanations, offered insights into individual predictions, ensuring coherence in the model's reasoning processes. 

No significant gender bias was found when exploring potential biases using partial dependence plots. However, a degree of bias was observed between initial eGFR values and CKD predictions. The system's logic was aligned with established biomedical literature, as evidenced in the biomedical relevance component, which extracted the logic through scoped rules and then cross-checked it with current knowledge. Safety assessments, in which the model was tested on potentially dangerous cases and the samples that it predicted incorrectly were analyzed, ensured the model's safety. 

The contribution of this research lies in its novel explainable system for CKD prediction, which transcends traditional feature importance analysis to offer a multifaceted view of the model. While tailored for CKD prediction, the system's framework harbors the potential for adaptation to other medical domains.

%Bibliography
\bibliographystyle{unsrt}  
\bibliography{ms}  

\end{document}